\documentclass[letterpaper]{article} 
\usepackage{aaai25}  
\usepackage{times}  
\usepackage{helvet}  
\usepackage{courier}  
\usepackage[hyphens]{url}  
\usepackage{graphicx} 
\usepackage{natbib}  
\usepackage{caption} 
\usepackage{algorithm}
\usepackage{algorithmic}

\usepackage{newfloat}
\usepackage{listings}
\DeclareCaptionStyle{ruled}{labelfont=normalfont,labelsep=colon,strut=off} 
\floatstyle{ruled}
\newfloat{listing}{tb}{lst}{}
\floatname{listing}{Listing}
\usepackage{multirow}
\usepackage{adjustbox}
\usepackage{amssymb}

\title{Mixture of Knowledge Minigraph Agents for Literature Review Generation}
\author {
    Zhi Zhang\textsuperscript{\rm 1},
    Yan Liu\textsuperscript{\rm 1,}\thanks{Corresponding author.},
    Sheng-hua Zhong\textsuperscript{\rm 2},
    Gong Chen\textsuperscript{\rm 1},
    Yu Yang\textsuperscript{\rm 3},
    Jiannong Cao\textsuperscript{\rm 1}
}
\affiliations {
    \textsuperscript{\rm 1}The Hong Kong Polytechnic University, the Department of Computing, Hong Kong, 999077, China\\
    \textsuperscript{\rm 2}Shenzhen University, the College of Computer Science and Software Engineering, Shenzhen, 518052, Guangdong, China\\
    \textsuperscript{\rm 3}The Education University of Hong Kong, Centre for Learning, Teaching, and Technology, Hong Kong, 999077, China\\
    zhi271.zhang@connect.polyu.hk, yan.liu@polyu.edu.hk, csshzhong@szu.edu.cn, csgchen@comp.polyu.edu.hk, yangyy@eduhk.hk, jiannong.cao@polyu.edu.hk
}

\usepackage{bibentry}

\begin{document}

\maketitle

\begin{abstract}
Literature reviews play a crucial role in scientific research for understanding the current state of research, identifying gaps, and guiding future studies on specific topics. However, the process of conducting a comprehensive literature review is yet time-consuming. This paper proposes a novel framework, collaborative knowledge minigraph agents (CKMAs), to automate scholarly literature reviews. A novel prompt-based algorithm, the knowledge minigraph construction agent (KMCA), is designed to identify relations between concepts from academic literature and automatically constructs knowledge minigraphs. By leveraging the capabilities of large language models on constructed knowledge minigraphs, the multiple path summarization agent (MPSA) efficiently organizes concepts and relations from different viewpoints to generate literature review paragraphs. We evaluate CKMAs on three benchmark datasets. Experimental results show the effectiveness of the proposed method, further revealing promising applications of LLMs in scientific research.
\end{abstract}

\begin{links}
    \link{Project}{https://minigraph-agents.github.io/}
\end{links}

\section{Introduction}

Artificial intelligence (AI) is being increasingly integrated into scientific discovery to augment and accelerate scientific research \cite{wang2023scientific}. Researchers are developing AI algorithms for various purposes, including literature understanding, experiment development, and manuscript draft writing \cite{liu2023generating,wang2024towards,martin2024shallow}.

Literature reviews play a crucial role in scientific research, assessing and integrating previous research on specific topics \cite{bolanos2024artificial}. They aim to meticulously identify and appraise all relevant literature related to a specific research question. Recent advancements in AI have shown promising performance in understanding research papers \cite{van2021automation}. By leveraging AI capabilities, automatic literature review algorithms enable researchers to save time and effort in the manual process of conducting literature reviews, rapidly identify key trends and gaps in recent research outputs, and uncover insights that might be overlooked in manual reviews \cite{wagner2022artificial}.

Automatic literature review algorithms typically involve two stages \cite{shi2023towards}: (1) selecting relevant reference documents and (2) summarizing the reference documents to compose a summary that presents the evolution of a specific field (these stages can be applied iteratively). Multiple scientific document summarization (MSDS), which aims to generate coherent and concise summaries for clusters of relevant scientific papers on a topic, is the representative work in the second stage. Over the past decades \cite{jin2020multi}, researchers have developed various summarization methods. Extractive methods directly select important sentences from original papers, while abstractive methods can generate new words and sentences but are technically more challenging than extractive methods.

\begin{figure}
    \begin{center}
        \includegraphics[width=0.9\linewidth]{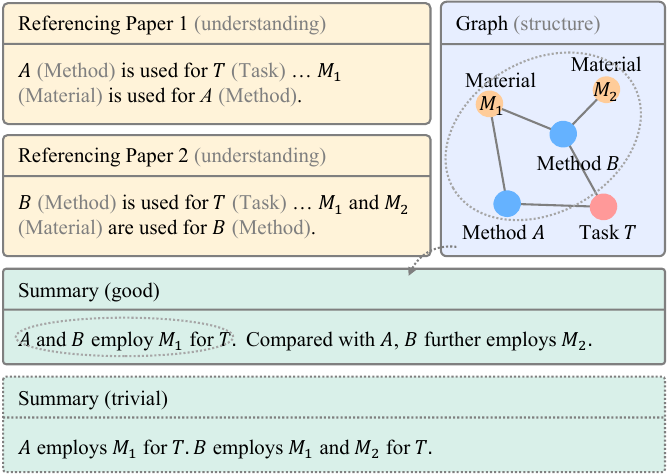}
    \end{center}
    \caption{Illustration of relations between concepts in literature review.}
    \label{fig:paradigm}
\end{figure}

Large Language Models (LLMs), pre-trained on massive text data, have shown human-like performance in language understanding and coherent synthesis, recently attracting interest in abstractive summarization. Though advanced in natural language processing, MSDS involves concepts that form complex relations, which LLMs are not naturally designed for and face challenges to organize. As shown in Fig. \ref{fig:paradigm}, reference documents involve multiple materials, methods, and tasks that are interconnected. A good summarization should organize concepts and their relations, merging consistent ones (e.g., $A$ and $B$ use $M_1$) and contrasting different ones (e.g., $B$ uses $M_2$ compared with $A$). Without explicit instructions, LLMs fail to model these relations and produce high-quality literature reviews \cite{li2024related}.

\begin{figure*}
    \begin{center}
        \includegraphics[width=\linewidth]{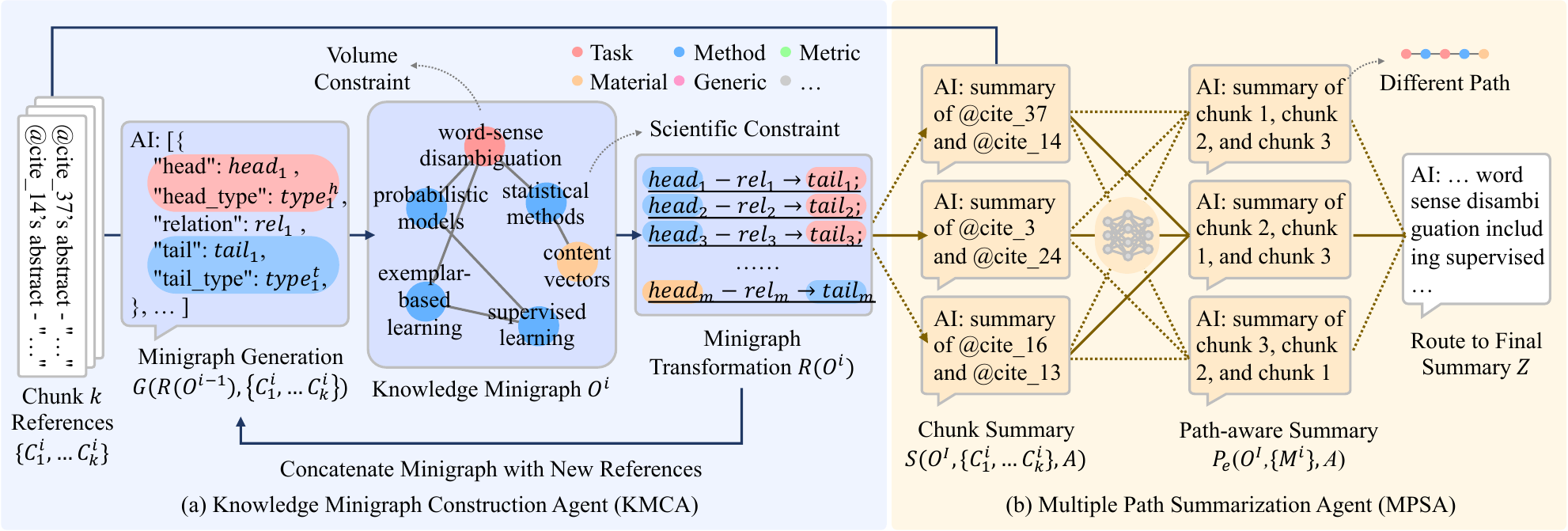}
    \end{center}
    \caption{The overall architecture of the proposed collaborative knowledge minigraph agents (CKMAs).}
    \label{fig:overview}
\end{figure*}

To handle these complex relations, we propose equipping LLMs with structural knowledge. Knowledge graphs, which represent entities as nodes and their relations as edges, are potential solutions. However, it is challenging to find a general-purpose knowledge graph that covers relations in various literature reviews. Instead of having a single graph for everything \cite{narayanan2017graph2vec}, we propose knowledge minigraphs for reference documents of interest. Knowledge minigraphs are small-scale graphs that comprise concepts dynamically extracted from reference documents as nodes and their relations as edges. They omit detailed documentation and highlight relations between concepts.

To automatically construct knowledge minigraphs, we propose a prompt-based algorithm, the knowledge minigraph construction agent (KMCA), to constrain LLMs in identifying research-relevant concepts and relations from references. To handle long context formed by reference documents, we design an iterative construction strategy, where key information and relations are iteratively extracted and stored from references into minigraphs.

By leveraging the knowledge minigraphs, the multiple path summarization agent (MPSA) is designed to organize the generated literature review. However, multiple valid viewpoints exist for discussing concepts through different paths in the knowledge minigraph. Thus, MPSA samples multiple summaries from different viewpoints in the knowledge minigraph, utilizing the technique of mixture of experts. A self-evaluation mechanism is then employed to automatically route to the most desirable summary as the final output.

\section{Related Work}

\subsection{Graphs in MSDS Tasks}

To generate a summary that is representative of the overall content, graph-based methods construct external graphs to assist document representation and cross-document relation modeling, achieving promising progress. In this regard, LexRank \cite{erkan2004lexrank} and TextRank \cite{mihalcea2004textrank} first introduce graphs to extractive text summarization in 2004. They compute sentence importance using a graph representation of sentences to extract salient textual units from documents as summarization. In 2020, Wang et al. propose to extract salient textual units from documents as summarization using a heterogeneous graph consisting of semantic nodes at several granularity levels of documents \cite{wang2020heterogeneous}. In 2022, Wang et al. incorporate knowledge graphs into document encoding and decoding, generating the summary from a knowledge graph template to achieve state-of-the-art performance \cite{wang2022multi}. However, to the best of our knowledge, no existing work integrates LLMs into graph-based methods to leverage their natural language understanding capabilities for improved graph construction and summary generation.

\subsection{Pre-trained Language Models in MSDS Tasks}

In recent years, pre-trained language models (PLMs) have demonstrated promising results in multiple document summarization. Liu et al. propose fine-tuning a pre-trained BERT model as the encoder and a randomly initialized decoder to enhance the quality of generated summaries \cite{liu2019text}. Xiao et al. introduce PRIMERA, a pre-trained encoder-decoder multi-document summarization model, by improving aggregating information across documents \cite{xiao2022primera}. More recently, pre-trained large language models (LLMs) show promising generation adaptability by training billions of model parameters on massive amounts of text data \cite{zhao2023survey,minaee2024large}. Zhang et al. utilize well-designed instructions to extract key elements, arrange key information, and generate summaries \cite{zhang20243a}. Zakkas et al. propose a three-step approach to select papers, perform single-document summarization, and aggregate results \cite{zakkas2024sumblogger}. PLMs can provide fluent summary results for literature review. However, they fall short of relation modeling in multiple reference documents. 

\section{Method}

Fig. \ref{fig:overview} illustrates the architecture of the proposed collaborative knowledge minigraph agents (CKMAs). CKMAs consist of two key components: the knowledge minigraph construction agent and the multiple path summarization agent.

\subsection{Knowledge Minigraph Construction Agent}

In this module, we are given $T$ reference documents $\{C_1, \ldots, C_T\}$'s abstracts. We aim to construct a knowledge structure that captures the relations between concepts in the referenced papers.

Past decades have witnessed knowledge graphs become the basis of information systems that require access to structured knowledge \cite{zou2020survey}. Knowledge structures are represented as semantic graphs, where nodes denote entities and are connected by relations denoted by edges. However, the general-purpose knowledge graphs are unsuitable for scientific document summarization, as they do not necessarily involve the main ideas of research papers. Thus, in this paper, we propose establishing a knowledge minigraph, defined as as a small set of research-relevant concepts and their relations. The construction steps of the knowledge minigraph are as follows:

\textbf{Reference chunking} Given a total of $T$ reference documents, we first divide them into $I$ chunks, each containing at most $k$ reference documents. Here, we iteratively select $k$ documents randomly to form each chunk, ensuring no document appears in multiple chunks until all documents are assigned. If $T$ is not perfectly divisible by $k$, the final chunk contains the remaining documents fewer than $k$. This chunking approach is necessary because MSDS usually involves numerous reference papers, forming a long context. LLMs either fail to process the entire context exceeding the acceptable length or suffer from missing crucial information positioned amidst a lengthy context \cite{zhang2024found}. Thus, we chunk related works and use LLMs to construct the knowledge structure step by step with $k$ reference papers each time. Here, $\{C_1^i, \ldots C_k^i\}$ represents the $k$ reference papers in the $i$-th chunk.  

\textbf{Minigraph generation} We employ a knowledge minigraph prompt $G(\{C_1^i, \ldots C_k^i\})$ to construct the knowledge structure of interest based on the abstracts of referenced papers $\{C_1^i, \ldots C_k^i\}$. It is known that constructing knowledge graphs from raw text data requires entity recognition and relation extraction \cite{trajanoska2023enhancing}. With recent advancements in prompt techniques, we design instructions with a demonstration to enable LLMs to perform these tasks within a single round of dialogue (using the prompt for query and LLMs for response). As shown in Table \ref{tab:prompt_table}, the prompt involves three special designs:

\begin{table*}
    \begin{center}
        \includegraphics[width=0.9\linewidth]{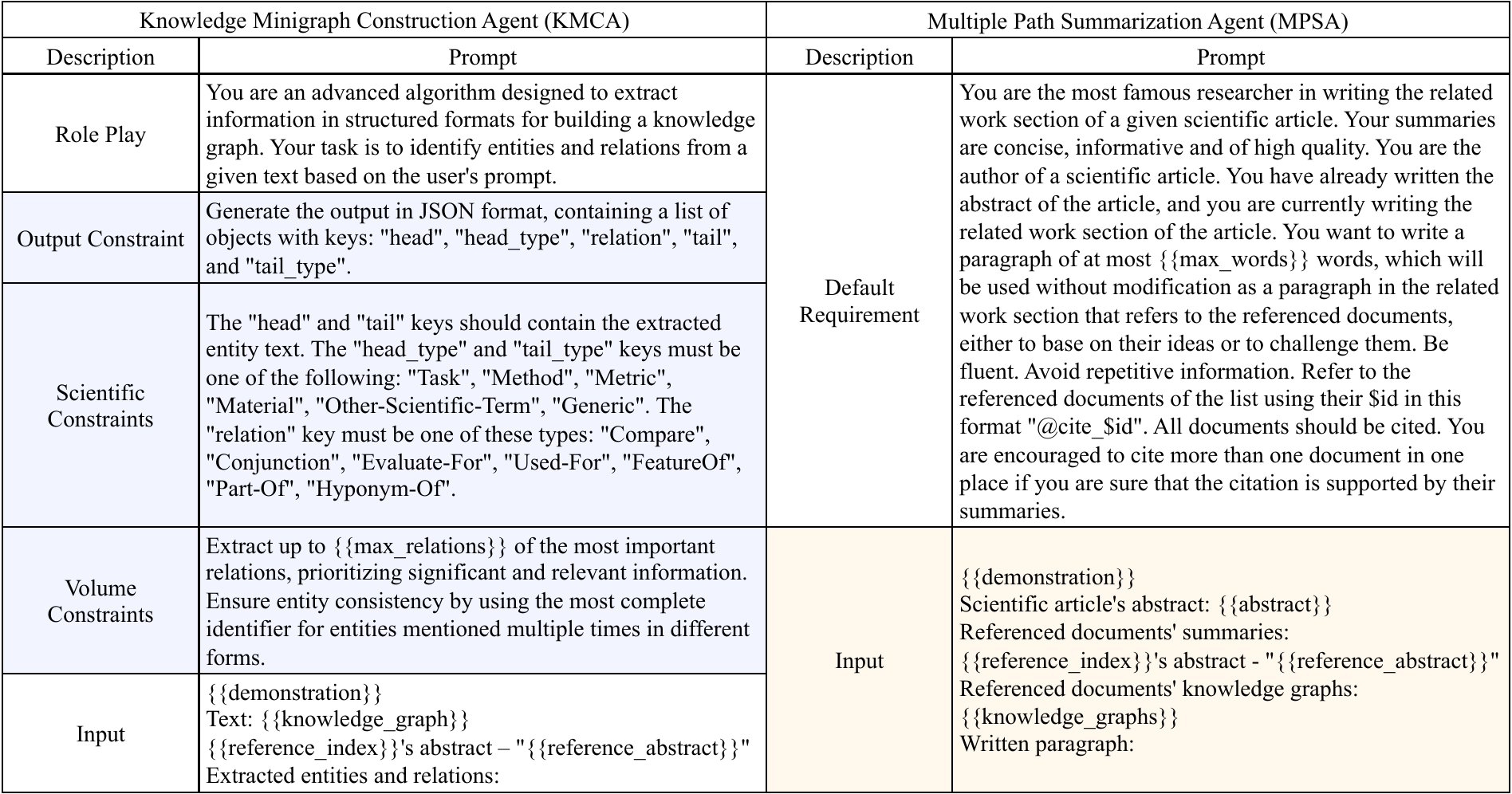}
    \end{center}
    \caption{Prompts in the knowledge minigraph construction agent and the multiple path summarization agent.}
  \label{tab:prompt_table}
\end{table*}

(1) Output constraint: LLMs are known for being relatively verbose and free-form in their output, making it hard for automated graph construction programs \cite{tan2023large}. Thus, we prompt LLM to constrain output in a machine-understandable JSON format. (2) Scientific constraints: to ensure the constructed knowledge structure revolves around the main idea of research topics, we design constraints on entities of interest and relations of interest. Inspired by DYGIE++ \cite{wadden2019entity}, we constrain entities to six types: Task, Method, Metric, Material, Generic, and OtherScientificTerm, and relations to seven types: Compare, Used-for, Feature-of, Hyponym-of, Evaluate-for, Part-of, and Conjunction. The definition of entities and relations for scientific knowledge graph construction has been studied in \cite{luan2018multi}. (3) Volume constraints: redundant relations will lead to long context exceeding LLM limits. To ensure the constructed knowledge structure is concise and informative, we constrain LLM to identify and select the $m$ most significant relations. Here, $m$ can be determined differently in applications. A small volume constraint may lead to meaningful relations outside the final knowledge minigraph. The maximum volume is bounded by the acceptable context length of LLMs.

\textbf{Minigraph transformation} To enable LLMs to understand the derived knowledge minigraph, we design a function $R(.)$ to transform the knowledge minigraph into a text representation for subsequent processing. In detail, available relations whose type meet the constraints are transformed into a line of text in the format $head_p - rel_p \rightarrow tail_p$, where $head_p$ and $tail_p$ are the head and tail entities, and $rel_p$ is the relation.

Finally, iteratively employing these three steps, the knowledge structure $O^i$ is constructed as shown in Eq. \ref{eq:dynamic_micro_ontology_generation}.
\begin{equation}
O^i=\left\{\begin{array}{l}
 G(R(O^{i-1}),\left\{C_1^i, \ldots C_k^i\right\}); i \geq 2 \\
 G(\left\{C_1^i, \ldots C_k^i\right\}) ; i=1
    \end{array}\right.
\label{eq:dynamic_micro_ontology_generation}
\end{equation}
In the first iteration, we apply the knowledge minigraph prompt $G$ to the first chunk of reference papers to derive the initial knowledge minigraph $O^1$. For subsequent iterations, we transform the intermediate knowledge minigraph $O^{i-1}$ into a text representation $R(O^{i-1})$. This text representation is then input along with the $i$-th chunk of reference papers $\left\{C_1^i, \ldots C_k^i\right\}$, to derive knowledge minigraph $O^i$. After $I$ iterations, we obtain the final knowledge minigraph $O^I$.

\subsection{Multiple Path Summarization Agent}

In this module, we are given the knowledge structure $O^I$, the referencing paper's abstract $A$, and the chunked referencing papers $\{C_1^i, \ldots C_k^i\}$'s abstracts. We aim to generate a summary following the knowledge structure.

Even given $O^I$ as guidance, generating a summary remains an ill-posed problem, i.e., the solution is not unique and depends on the specific discussion viewpoints. For example, one can highlight different concepts or choose different writing logic for various situations. How do we handle varying viewpoints of the knowledge structure and create a more capable summary? Inspired by the mixture of experts approach \cite{shazeer2017outrageously}, a machine learning technique to leverage diverse model capabilities where multiple expert networks specialize in different skill sets, we propose using LLMs with different hinted paths to understand the knowledge minigraph for generating multiple summaries and selecting the best viewpoint. The steps of the multiple path summarization agent are as follows:

\textbf{Chunk summarization} As mentioned before, MSDS usually involves numerous reference papers, forming a long context problem. We chunk them into $I$ chunks and formulate a hierarchical summarization process, first generating summaries for each chunk of referenced papers. With prompt engineering, we elicit the behavior of LLMs to create summaries for each chunk under the guidance of derived knowledge minigraphs $O^I$. As illustrated in Table \ref{tab:prompt_table}, we instruct LLMs to take into consideration three kinds of information: the scientific article's abstract $A$, summaries of referenced paper $\{C_1^i, \ldots C_k^i\}$, and the knowledge minigraphs of referenced paper $O^I$. We provide a demonstration to help LLMs understand the task. Then, LLMs respond to write a summary for each chunk, where $A$ and $\{C_1^i, \ldots C_k^i\}$ provide textual details locally and $O^I$ provide structural knowledge globally. The instructions are revised based on related work for fair comparison \cite{zakkas2024sumblogger}. In real-world applications, users can customize these instructions, such as specifying writing style. Mathematically, the chunk summarization can be denoted as:
\begin{equation}
    M^i = S(A,\{C_1^i, \ldots C_k^i\}, O^I)
    \label{eq:chunk_summary}
\end{equation}
where $S$ is the prompt function for chunk summarization.

\textbf{Path-aware Summarization} We employ $E$ experts to merge all chunk summaries ${M^i}$ and generate final summaries, with each expert aware of different hinted paths to understand the knowledge minigraph. Given consistent knowledge structure, different human researchers may have varying interpretations, selectively emphasizing concepts in a rational order. To automatically mimic human researchers and generate summaries reflecting different understandings, we leverage the observation that LLMs are sensitive to the order of prompt wording \cite{pezeshkpour2023large}. We find that the order of given reference papers impacts the generated summary, affecting the organization of concepts (e.g., the order of introducing concepts). Thus, we apply simple random sampling to obtain $E$ permutations from the full permutations of reference papers to serve as hints of potential paths in the knowledge minigraph. These sampled permutations are then used to prompt the LLM to generate summaries. We use the same prompt as in chunk summarization, except that $\{C_1^i, \ldots, C_k^i\}$ are replaced by a permutation of chunk summaries ${M^i}$. Mathematically, the summaries generated by the $e$-th expert can be denoted as:
\begin{equation}
Y_e=P_e(A,\{M^i\}, O^I)
\label{eq:expert_summary}
\end{equation}
where $P_e$ represents the prompt function for the $e$-th expert with the sampled $e$-th permutation of referencing papers. For instance, given chunk summarizations $\{M_1,M_2,M_3\}$, where $M_1$, $M_2$, and $M_3$ are summaries of the first, second, and third chunks of referencing papers, respectively, three experts can be fed $[M_1,M_2,M_3]$, $[M_2,M_1,M_3]$, and $[M_3,M_2,M_1]$ as input. The remaining parts of the instructions remain consistent with $S$.

\begin{table*}
    \begin{center}
        \begin{tabular}{|ll|rr|rr|rr|}
            \hline
            \multicolumn{2}{|c|}{\multirow{2}{*}{Method}}                            & \multicolumn{2}{c|}{Multi-Xscience}                 & \multicolumn{2}{c|}{TAD}                            & \multicolumn{2}{c|}{TAS2}                           \\ \cline{3-8} 
            \multicolumn{2}{|c|}{}                                                   & \multicolumn{1}{c|}{ROUGE-1}        & ROUGE-2       & \multicolumn{1}{c|}{ROUGE-1}        & ROUGE-2       & \multicolumn{1}{c|}{ROUGE-1}        & ROUGE-2       \\ \hline
            \multicolumn{1}{|c|}{\multirow{6}{*}{Graphs}} & LexRank (2004)           & \multicolumn{1}{r|}{30.19}          & 5.53          & \multicolumn{1}{r|}{27.29}          & 3.50          & \multicolumn{1}{r|}{27.04}          & 3.18          \\ \cline{2-8} 
            \multicolumn{1}{|c|}{}                        & TexRank (2004)           & \multicolumn{1}{r|}{31.51}          & 5.83          & \multicolumn{1}{r|}{26.80}          & 3.61          & \multicolumn{1}{r|}{26.19}          & 3.14          \\ \cline{2-8} 
            \multicolumn{1}{|c|}{}                        & HeterSumGraph (2020)     & \multicolumn{1}{r|}{31.36}          & 5.82          & \multicolumn{1}{r|}{27.85}          & 3.88          & \multicolumn{1}{r|}{27.56}          & 3.62          \\ \cline{2-8} 
            \multicolumn{1}{|c|}{}                        & GraphSum (2020)          & \multicolumn{1}{r|}{29.58}          & 5.54          & \multicolumn{1}{r|}{26.12}          & 4.03          & \multicolumn{1}{r|}{25.01}          & 3.23          \\ \cline{2-8} 
            \multicolumn{1}{|c|}{}                        & TAG (2022)               & \multicolumn{1}{r|}{33.45}          & 7.06          & \multicolumn{1}{r|}{30.48}          & 6.16          & \multicolumn{1}{r|}{28.04}          & 4.75          \\ \cline{2-8} 
            \multicolumn{1}{|c|}{}                        & KGSum (2022)             & \multicolumn{1}{r|}{35.77}          & 7.51          & \multicolumn{1}{r|}{32.38}          & 5.19          & \multicolumn{1}{r|}{30.67}          & 4.76          \\ \hline
            \multicolumn{1}{|c|}{\multirow{11}{*}{PLMs}}  & Pointer-Generator (2017) & \multicolumn{1}{r|}{34.11}          & 6.76          & \multicolumn{1}{r|}{31.70}          & 6.41          & \multicolumn{1}{r|}{28.53}          & 4.96          \\ \cline{2-8} 
            \multicolumn{1}{|c|}{}                        & BertABS (2019)           & \multicolumn{1}{r|}{31.56}          & 5.02          & \multicolumn{1}{r|}{27.42}          & 4.88          & \multicolumn{1}{r|}{25.45}          & 3.82          \\ \cline{2-8} 
            \multicolumn{1}{|c|}{}                        & SciBertABS (2019)        & \multicolumn{1}{r|}{32.12}          & 5.59          & \multicolumn{1}{r|}{27.88}          & 5.19          & \multicolumn{1}{r|}{26.01}          & 4.13          \\ \cline{2-8} 
            \multicolumn{1}{|c|}{}                        & HiMAP (2019)             & \multicolumn{1}{r|}{31.66}          & 5.91          & \multicolumn{1}{r|}{30.49}          & 6.21          & \multicolumn{1}{r|}{28.37}          & 5.07          \\ \cline{2-8} 
            \multicolumn{1}{|c|}{}                        & BART (2020)              & \multicolumn{1}{r|}{32.83}          & 6.36          & \multicolumn{1}{r|}{25.39}          & 4.74          & \multicolumn{1}{r|}{27.73}          & 4.80          \\ \cline{2-8} 
            \multicolumn{1}{|c|}{}                        & MGSum (2020)             & \multicolumn{1}{r|}{33.11}          & 6.75          & \multicolumn{1}{r|}{27.49}          & 4.79          & \multicolumn{1}{r|}{25.54}          & 3.75          \\ \cline{2-8} 
            \multicolumn{1}{|c|}{}                        & PRIMERA (2022)           & \multicolumn{1}{r|}{31.90}          & 7.40          & \multicolumn{1}{r|}{32.04}          & 5.78          & \multicolumn{1}{r|}{29.99}          & 5.07          \\ \cline{2-8} 
            \multicolumn{1}{|c|}{}                        & GPT-3.5-turbo (2023)     & \multicolumn{1}{r|}{31.11}          & 7.38          & \multicolumn{1}{r|}{30.77}          & 4.78          & \multicolumn{1}{r|}{26.97}         & 4.14          \\ \cline{2-8} 
            \multicolumn{1}{|c|}{}                        & GPT-4 (2023)             & \multicolumn{1}{r|}{33.21}          & 7.61          & \multicolumn{1}{r|}{32.50}          & 4.90          & \multicolumn{1}{r|}{30.71}          & 4.25          \\ \cline{2-8} 
            \multicolumn{1}{|c|}{}                        & 3A-COT (2024)            & \multicolumn{1}{r|}{23.65}          & 4.85          & \multicolumn{1}{r|}{23.02}          & 3.73          & \multicolumn{1}{r|}{22.65}          & 3.43          \\ \cline{2-8} 
            \multicolumn{1}{|c|}{}                        & SumBlogger (2024)        & \multicolumn{1}{r|}{35.40}          & 8.40          & \multicolumn{1}{r|}{33.90}          & 5.51          & \multicolumn{1}{r|}{30.69}          & 3.92          \\ \hline
            \multicolumn{2}{|c|}{Proposed}                                           & \multicolumn{1}{r|}{\textbf{36.41}} & \textbf{8.78} & \multicolumn{1}{r|}{\textbf{34.16}} & \textbf{6.22} & \multicolumn{1}{r|}{\textbf{32.31}} & \textbf{5.36} \\ \hline
        \end{tabular}
    \end{center}
    \caption{Comparison of CKMAs with state-of-the-art methods on Multi-Xscience, TAS2, and TAD datasets. A higher ROUGE score indicates better performance. The best results are highlighted in bold.}
\label{tab:comparison_experiments}
\end{table*}

\textbf{Summarization router} We design a router to evaluate different experts' summaries and automatically select the most desirable summary $Y_e$ as the final output. Without requiring additional side information, this paper proposes a self-evaluation strategy. In detail, we observe that there are agreements between the experts' viewpoints and their generated summaries. We propose to quantify the degree of agreement for each summary using the ROUGE-1 score \cite{lin2004rouge}, which measures the overlap between a generated summary and other summaries. We then select the summary with the highest degree of agreement, which indicates that its understanding has the highest likelihood of being supported by other experts, or in other words, is relatively more acceptable. Mathematically, the final summary $Z = Y_{e^*}$ is determined by:

\begin{equation}
    e^* = \arg \max_e \sum_{j \neq e} \mathrm{rouge1}(Y_e, Y_j)
    \label{eq:final_summary}
\end{equation}
where $\mathrm{rouge1}(Y_e, Y_j)$ is the an 1-gram recall (ROUGE-1 score) between $e$-expert's generated summary $Y_e$ and $j$-expert's generated summary $Y_j$.

\section{Experiments}

We evaluate our approach on three public MSDS datasets: Multi-Xscience \cite{lu2020multi}, TAD \cite{chen2022target}, and TAS2 \cite{chen2022target}. All datasets share a consistent format for input and ground-truth pairs: each sample comprises a query paper's abstract and its cited reference abstracts as input, with a related work paragraph from the query paper serving as the ground truth summary. We use GPT-3.5-turbo as the backbone model with a temperature set to 0.0 for reproducibility. We set the chunk size $k$ to 3 and the number of experts $E$ to 3. We set the volume constraint $m$ to 32. Following previous work, we automatically evaluate the summarization quality using ROUGE scores \cite{lin2004rouge}. We employ ROUGE-N to calculate the N-grams overlap between the output and gold summary, measuring the summary quality. We report ROUGE-1 and ROUGE-2 metrics for unigram and bigram co-occurrences.

\begin{table*}
    \begin{center}
        \begin{adjustbox}{width=\linewidth}
            \begin{tabular}{|cccrr|cccrr|}
                \hline
                \multicolumn{5}{|c|}{Knowledge Minigraph Construction Agent (KMCA)}                                                                                                                            & \multicolumn{5}{c|}{Multiple Path Summarization Agent (MPSA)}                                                                                                                                      \\ \hline
                \multicolumn{1}{|c|}{Scientific} & \multicolumn{1}{c|}{Volume}     & \multicolumn{1}{c|}{Iterative}      & \multicolumn{1}{c|}{\multirow{2}{*}{ROUGE-1}} & \multirow{2}{*}{ROUGE-2} & \multicolumn{1}{c|}{Chunk}   & \multicolumn{1}{c|}{Path} & \multicolumn{1}{c|}{Summary}  & \multicolumn{1}{c|}{\multirow{2}{*}{ROUGE-1}} & \multirow{2}{*}{ROUGE-2} \\
                \multicolumn{1}{|l|}{Constraint} & \multicolumn{1}{l|}{Constraint} & \multicolumn{1}{l|}{Construction} & \multicolumn{1}{c|}{}                         &                          & \multicolumn{1}{l|}{Summary} & \multicolumn{1}{c|}{Permutation}                             & \multicolumn{1}{c|}{Router} & \multicolumn{1}{c|}{}                         &                          \\ \hline
                \multicolumn{1}{|c|}{$\times$}          & \multicolumn{1}{c|}{$\times$}          & \multicolumn{1}{c|}{$\times$}            & \multicolumn{1}{r|}{34.90}                    & 8.56                     & \multicolumn{1}{c|}{$\times$}       & \multicolumn{1}{c|}{$\times$}                            & \multicolumn{1}{c|}{$\times$}      & \multicolumn{1}{r|}{32.04}                    & 5.54                     \\ \hline
                \multicolumn{1}{|c|}{$\times$}          & \multicolumn{1}{c|}{$\checkmark$}          & \multicolumn{1}{c|}{$\checkmark$}            & \multicolumn{1}{r|}{35.69}                    & 8.62                     & \multicolumn{1}{c|}{$\times$}       & \multicolumn{1}{c|}{$\checkmark$}                            & \multicolumn{1}{c|}{$\checkmark$}      & \multicolumn{1}{r|}{33.29}                    & 6.47                     \\ \hline
                \multicolumn{1}{|c|}{$\checkmark$}          & \multicolumn{1}{c|}{$\times$}          & \multicolumn{1}{c|}{$\checkmark$}            & \multicolumn{1}{r|}{35.50}                    & 8.59                     & \multicolumn{1}{c|}{$\checkmark$}       & \multicolumn{1}{c|}{$\times$}                            & \multicolumn{1}{c|}{$\checkmark$}      & \multicolumn{1}{r|}{34.00}                    & 7.09                     \\ \hline
                \multicolumn{1}{|c|}{$\checkmark$}          & \multicolumn{1}{c|}{$\checkmark$}          & \multicolumn{1}{c|}{$\times$}            & \multicolumn{1}{r|}{35.04}                    & 8.57                     & \multicolumn{1}{c|}{$\checkmark$}       & \multicolumn{1}{c|}{$\checkmark$}                            & \multicolumn{1}{c|}{$\times$}      & \multicolumn{1}{r|}{32.18}                    & 6.32                     \\ \hline
                \multicolumn{1}{|c|}{$\checkmark$}          & \multicolumn{1}{c|}{$\checkmark$}          & \multicolumn{1}{c|}{$\checkmark$}            & \multicolumn{1}{r|}{\textbf{36.41}}           & \textbf{8.78}            & \multicolumn{1}{c|}{$\checkmark$}       & \multicolumn{1}{c|}{$\checkmark$}                            & \multicolumn{1}{c|}{$\checkmark$}      & \multicolumn{1}{r|}{\textbf{36.41}}           & \textbf{8.78}            \\ \hline
                \end{tabular}
        \end{adjustbox}
    \end{center}
    \caption{Ablation study of the proposed collaborative knowledge minigraph agents (CKMAs) on the Multi-Xscience dataset. A higher ROUGE score indicates better performance. The best results are highlighted in bold.}
\label{tab:ablation_experiments}
\end{table*}

\subsection{Comparison Experiments}

Table \ref{tab:comparison_experiments} compares the proposed model with graph-based methods including LexRank \cite{erkan2004lexrank}, TextRank \cite{mihalcea2004textrank}, HeterSumGraph \cite{wang2020heterogeneous}, GraphSum \cite{li2020leveraging}, TAG \cite{chen2022target}, and KGSum \cite{wang2022multi} and pre-trained language model-based
methods, including Pointer-Generator \cite{see2017get}, BertABS \cite{liu2019text}, SciBertABS \cite{beltagy2019scibert}, HiMAP \cite{fabbri2019multi}, BART \cite{lewis2020bart}, MGSum \cite{jin2020multi}, PRIMERA \cite{xiao2022primera}, GPT-3.5-turbo \cite{ouyang2022training}, and GPT-4 \cite{achiam2023gpt}. The proposed CKMAs achieve state-of-the-art performance on all three datasets in terms of ROUGE-1 and ROUGE-2 scores. CKMAs also outperform the latest prompt-powered MSDS, e.g., 3A-COT \cite{zhang20243a} and SumBlogger \cite{zakkas2024sumblogger}.

\subsection{Ablation Studies}

This section presents ablation studies to investigate the performance gains brought by the designed modules in CKMAs. We first ablate the knowledge minigraph construction agent (KMCA) and the multiple path summarization agent (MPSA), respectively. When ablating KMCA, we no longer construct the knowledge minigraph and do not include it as part of the MPSA instruction. When removing MPSA, we use the instruction in Table \ref{tab:prompt_table} to generate a single summary as the final summary directly. We report the performance of the ablated version in the first line of Table \ref{tab:ablation_experiments}.

Then, we ablate the modules in MPSA and KMCA. For the knowledge minigraph construction agent, when ablating the scientific constraint or volume constraint ablation in minigraph generation, we remove the corresponding instruction in Table \ref{tab:prompt_table}. To ablate iterative construction, we remove Eq. \ref{eq:dynamic_micro_ontology_generation} and directly use all references as input. Due to the length issue, the over-length context is truncated. For the multiple path summarization agent, when ablating chunk summarization, we directly use all references' abstracts as input. To ablate the path-aware summarization strategy, we use the references with original order as input, adjusting temperatures from 0.0 to 0.7. When removing the summarization router, we randomly select a generated summary.

\begin{figure}
    \begin{center}
        \includegraphics[width=\linewidth]{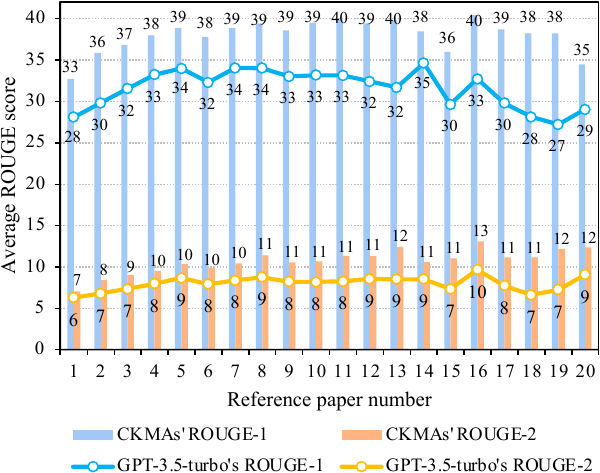}
    \end{center}
    \caption{ROUGE score comparison of the proposed CKMAs with its backbone models (GPT-3.5-turbo) group by reference paper number on the Multi-Xscience dataset.}
    \label{fig:rouge_scores}
\end{figure}

Table \ref{tab:ablation_experiments} shows the results of ablation studies, with the full version of the proposed model reported in the last line. We find that removing any module leads to performance degradation. This indicates that all designs contribute to the final performance. The MPSA brings a 4\% performance gain, and KMCA brings a 2\% performance gain. For designs in the knowledge minigraph construction agent, the performance gain brought by iterative construction is the largest, indicating its effectiveness in understanding long contexts. For designs in the multiple path summarization agent, the performance gain brought by the summarization router is the largest, indicating the importance of selecting the most desirable summaries from rational paths.

\subsection{Case Studies}

This section conducts case studies to provide further insights into our model's performance. We first perform statistical analysis to validate in which cases the model succeeds and in which it fails. We group the test samples based on the number of references in the gold summary. We calculate the average ROUGE-1 and ROUGE-2 scores for CKMAs and its backbone model (GPT-3.5-turbo) in every group, comparing generated and gold summaries. The results are presented in Fig. \ref{fig:rouge_scores}. We observe that CKMAs consistently outperform GPT-3.5-turbo regardless of the number of referencing papers. As the number of reference papers increases, the performance gap between CKMAs and GPT-3.5-turbo widens. It demonstrates the capability to model complex relations within a long context, in contrast to the performance decrease observed in GPT-3.5-turbo. For error analysis, the results demonstrate modest improvement when the number of reference papers is small. This is attributed to the limited relations in limited documents.

\begin{table*}
    \begin{center}
        \includegraphics[width=\linewidth]{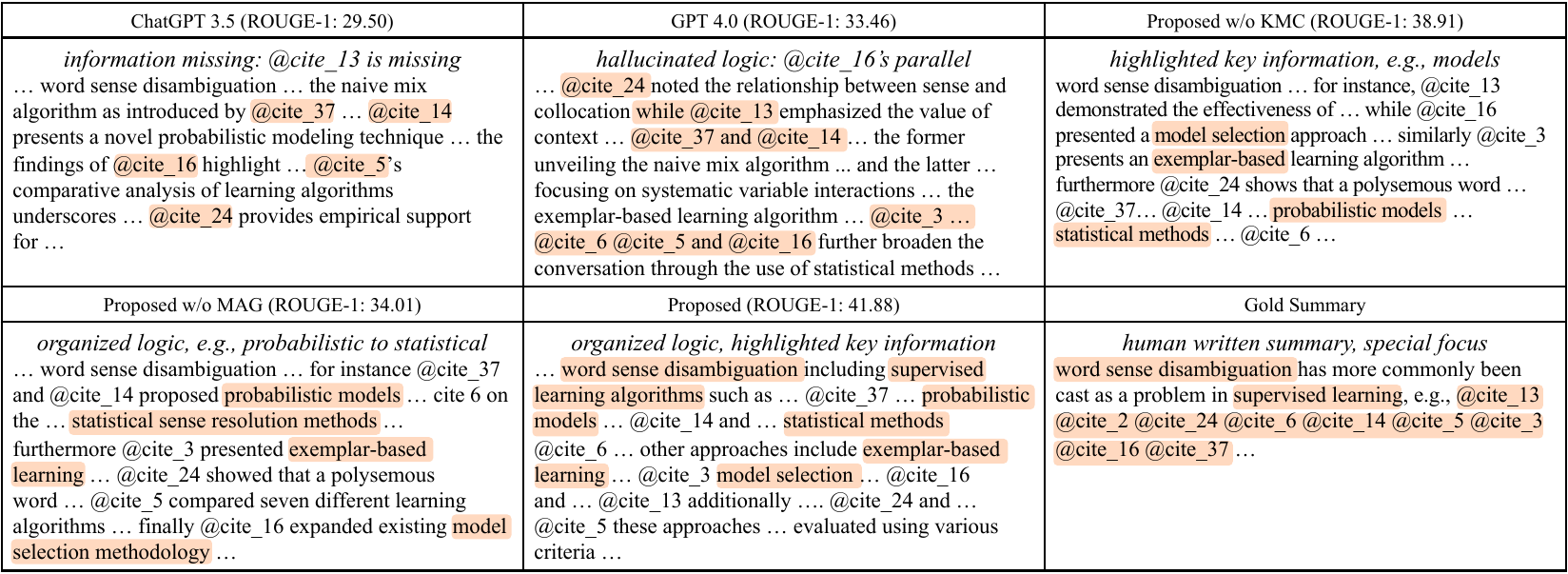}
    \end{center}
    \caption{Case study of the proposed collaborative knowledge minigraph agents (CKMAs) on the Multi-Xscience dataset.}
  \label{tab:prompt_example}
\end{table*}

For further insights, we sample an instance from the Multi-Xscience dataset and use well-known LLMs, GPT-3.5-turbo and GPT 4.0, to generate summaries. The generated results are shown in Table \ref{tab:prompt_example}. We find that GPT-3.5-turbo suffers from information loss, omitting citation 13. GPT 4.0 shows improvement but lists facts in parallel without rational connections. For example, citations 37 and 14 are listed side by side, but show no parallel relationship. We then use different versions of the proposed CKMAs to generate a summary. It can be observed that without the KMCA, the MPSA contributes to highlighting different categories of algorithms from the desired viewpoint. Without MPSA, the KMCA contributes to organizing algorithms rationally, e.g., from probabilistic to statistical approaches and then to example-based learning methods. With both modules, CKMAs generate the best summary.

\begin{figure}[t]
    \begin{center}
        \includegraphics[width=\linewidth]{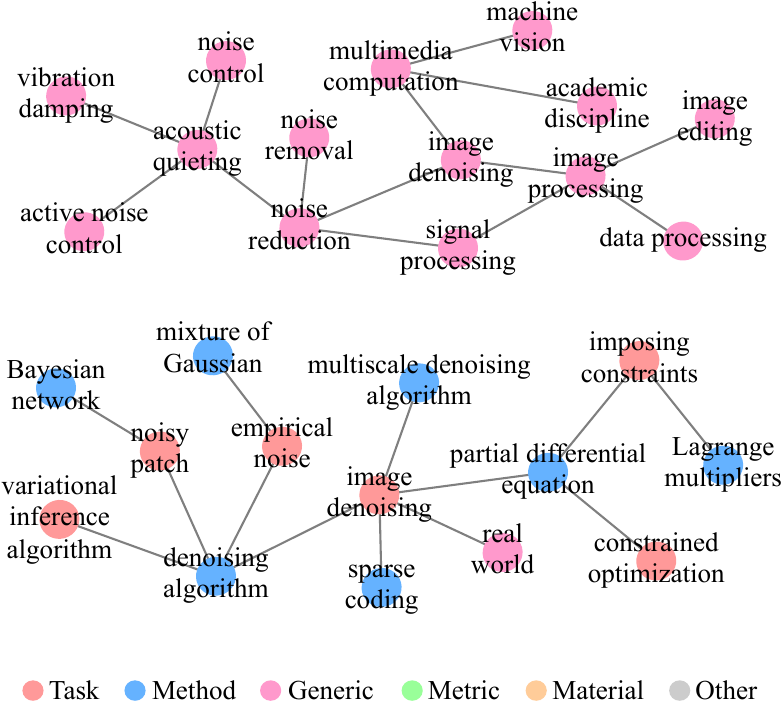}
    \end{center}
    \caption{Case study of knowledge graphs queried from Wikidata (top) and knowledge minigraph constructed by CKMAs (bottom) for the topic ``image denoising''.}
    \label{fig:kg_vs_mini_kg}
\end{figure}

We then analyze the differences between queried public knowledge graphs and the constructed knowledge minigraphs. We sample an instance from the Multi-Xscience dataset for this comparison. To query the knowledge graph, we use SPARQL to access Wikidata, a collaborative knowledge base. The queried knowledge graph is shown in the upper part of Fig. \ref{fig:kg_vs_mini_kg}. For the knowledge minigraph, we employ the proposed method with knowledge minigraph construction, with the result displayed in the lower part. We observe that the entities in the queried knowledge graph are general-purpose and lack specific insights into research problems. The minigraph clearly presents tasks and methods. It demonstrates GPT-3.5-turbo's reliability in knowledge minigraph construction, as supported by related studies \cite{zhu2024llms}. As a result, knowledge minigraphs provide more topic-focused information that enhances summarization.

\begin{figure}
    \begin{center}
        \includegraphics[width=0.95\linewidth]{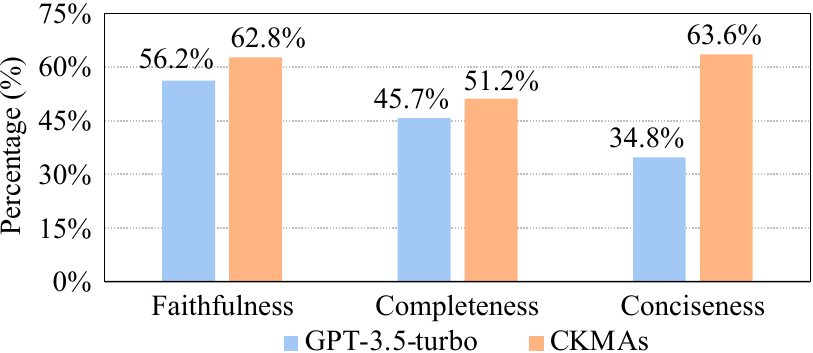}
    \end{center}
    \caption{Performance comparison of CKMAs with its backbone model (GPT-3.5-turbo) along faithfulness, completeness, and conciseness dimensions. A higher score indicates better performance.}
    \label{fig:finesure_scores}
\end{figure}

\subsection{Disscussion}

To gain further insights, besides similarity-based metrics (ROUGE), we incorporate FineSurE \cite{song2024finesure}, a recent LLM-based automated evaluation framework that assesses summarization quality across three dimensions: faithfulness (minimizing factual errors), completeness (covering key facts), and conciseness (avoiding redundant details).

We evaluated the proposed CKMAs against its backbone model on the Multi-Xscience dataset. Fig. \ref{fig:finesure_scores} demonstrates that CKMAs outperform GPT-3.5-turbo in conciseness and faithfulness, indicating that knowledge minigraphs effectively filter irrelevant relations while retaining viewpoints supported by related work. Both approaches exhibit moderate completeness scores, which can be attributed to the divergent purposes in individual review paragraphs (e.g., discussing feasibility versus discussing limitations).

\section{Conclusions and Future Work}

This paper aims to provide an intelligent research copilot to assist in writing literature reviews based on given references. While recent LLMs excel at natural language understanding and generation, they struggle to explicitly model complex relations in multiple reference documents. To address this challenge, we propose collaborative knowledge minigraph agents (CKMAs).

We conduct comparison experiments, ablation studies, and case studies on benchmark datasets. Experimental results show the effectiveness of the CKMAs, further revealing promising applications of LLMs in scientific research.

\section*{Acknowledgements}

This work was supported by the ``A Large Language Model-powered Clinical Communication Training System Under Cantonese-English Code-switching Scenario'' (ITS/397/23), the Hong Kong Research Grants Council (RGC) Theme-based Research Scheme (T43-513/23-N, T41-603/20-R), and the National Natural Science Foundation of China (62472291), with research partially conducted at the Research Institute for Artificial Intelligence of Things (RIAIoT) at PolyU.

\bibliography{aaai25}

\end{document}